\documentclass{article}

\PassOptionsToPackage{numbers, compress}{natbib}

\usepackage[final]{neurips_2021}

\usepackage{array}

\usepackage{algorithm}
\usepackage{algpseudocode}
\usepackage[utf8]{inputenc} 
\usepackage[T1]{fontenc}    
\usepackage{subcaption}
\usepackage{colortbl}
\usepackage{hyperref}       
\usepackage{url}            
\usepackage{booktabs}       
\usepackage{bbm}
\usepackage{amsfonts}       
\usepackage{wrapfig}
\usepackage{graphicx}
\usepackage{amsmath}       
\usepackage{xcolor}
\definecolor{darkblue}{HTML}{00008B}
\definecolor{darkred}{HTML}{8B0000}
\definecolor{darkgreen}{HTML}{006400}
\definecolor{darkpurple}{HTML}{663399}

\definecolor{darkyellow}{HTML}{d3a809}
\definecolor{darkorange}{HTML}{ff8c00}
\definecolor{darkmagenta}{HTML}{8b008b}
\usepackage{amssymb}       
\usepackage{nicefrac}       
\usepackage{microtype}      
\usepackage{mathtools}

\newcommand{\triple}[3]{\textcolor{darkred}{#1}/\textcolor{darkblue}{#2}/\textcolor{darkgreen}{#3}}

\newif\ifcomments
\commentsfalse
\newcommand\graycell{\cellcolor[rgb]{0.9,0.9,0.9}}
\newcommand\yellowcell{\cellcolor[rgb]{1.0,1.0,0.85}}
\newcommand\orangecell{\cellcolor[rgb]{1.0,0.92,0.88}}
\newcommand\magentacell{\cellcolor[rgb]{1.0,0.85,1.0}}

\DeclareMathOperator{\AMI}{AMI}
\DeclareMathOperator{\proj}{proj}
\DeclareMathOperator{\AND}{AND}
\DeclareMathOperator{\EditDistance}{EditDistance}

\title{Emergent Communication of Generalizations}

\author{%
  Jesse Mu \\
  Stanford University \\
  \texttt{muj@stanford.edu} \\
  \And
  Noah Goodman \\
  Stanford University \\
  \texttt{ngoodman@stanford.edu} \\
}

\begin{document}
\normalsize

\maketitle

\begin{abstract}
To build agents that can collaborate effectively with others, recent research has trained artificial agents to communicate with each other in Lewis-style referential games. However, this often leads to successful but uninterpretable communication. We argue that this is due to the game objective: communicating about a single object in a shared visual context is prone to overfitting and does not encourage language useful beyond concrete reference. In contrast, human language conveys a rich variety of abstract ideas. To promote such skills, we propose games that require communicating generalizations over \emph{sets} of objects representing abstract visual concepts, optionally with separate contexts for each agent. We find that these games greatly improve systematicity and interpretability of the learned languages, according to several metrics in the literature. Finally, we propose a method for identifying logical operations embedded in the emergent languages by learning an approximate compositional reconstruction of the language.
\end{abstract}

\section{Introduction}

\begin{figure}[t]
    \centering
    \includegraphics[width=\linewidth]{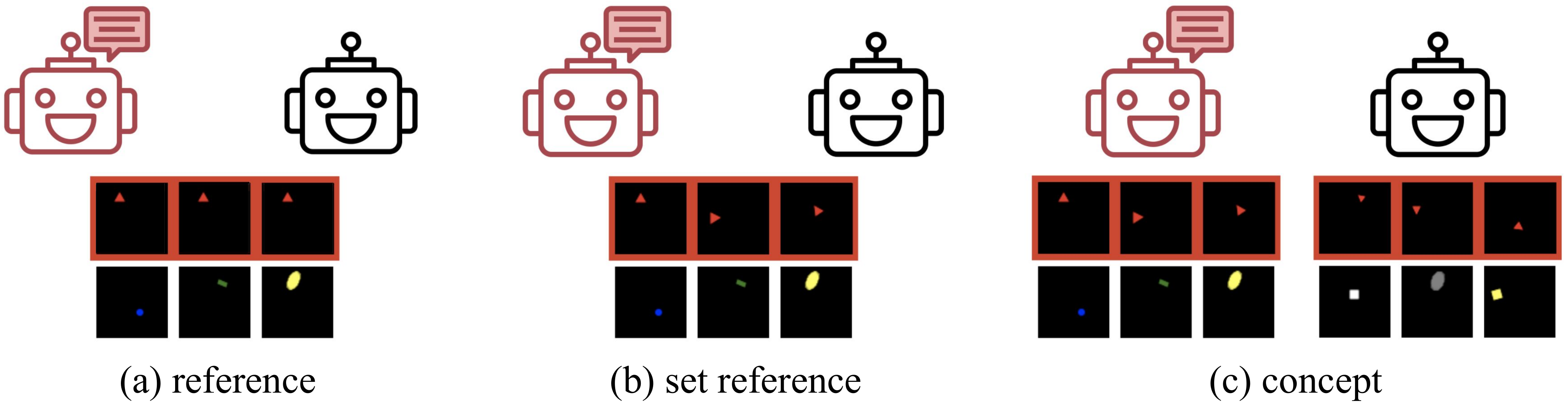}
    \caption{Communication games for the concept \emph{red triangle}. Given a set of targets (red borders) and distractors, a {\color{darkred} \textbf{teacher}} must send a message to help a \textbf{student} identify the targets. In (a) reference games, targets are identical; in (b) set reference (setref) games, there are multiple targets; and in (c) concept games, the agents see different inputs.}
    \label{fig:overview}
    \vspace{-1em}
\end{figure}

The communication systems that emerge when two agents are trained to cooperate offer a window on the evolution of human language, as well as a promising avenue for improving the collaboration abilities of artificial agents.
Much recent work studies Lewis-style \cite{Lewis1969} signaling games (Figure~\ref{fig:overview}a), where agents are trained to refer to a single object in a shared visual context. However, a general consensus of this work is that without careful environmental pressures, agents develop successful but uninterpretable communication schemes distinctly unlike human language \cite{andreas2019measuring,chaabouni2019anti,chaabouni2020compositionality,kottur2017natural,lazaridou2020emergent}.

We argue that the reference games typically used in these studies are ill-suited to drive linguistic systematicity for two reasons. One is perceptual: agents can exploit inscrutable patterns in single inputs, which leads to communication via spurious features \cite{bouchacourt2018agents}. The other reason is cognitive: human language can convey abstract ideas, such as kinds and causes, not only reference to specific objects. Simple reference games are unlikely to drive emergence of such abstract language.
In particular, \emph{generalizations} over categories are a crucial part of language \cite{tessler2019language}, helping us transfer knowledge that may be useful in the future. For example, we would like to teach our kin not just to avoid one specific lion, but to avoid all lions, including those that have not yet been seen. Some have even argued that language emerged precisely from this need to teach hard-won generalizations to others \cite{laland2017origins}. With this idea in mind, can we design an experimental setting that better catalyzes these abilities?

In this paper, we propose extensions of Lewis-style signaling games to \emph{sets}. In the \emph{set reference} (setref) game, a teacher must communicate to a student not just a single object, but rather a group of objects belonging to a concept (Figure~\ref{fig:overview}b). In the \emph{concept} game, each agent sees different examples of the concept  (Figure~\ref{fig:overview}c). Inspired by human teaching \cite{chopra2019ratchet}, our core insight is that requiring generalization to combinatorially large sets of (possibly unseen) objects encourages agents to learn and communicate rich abstractions across inputs (e.g.\ \emph{seagulls}), instead of low-level features (e.g.\ \emph{color \#FDA448}).
These tasks are more difficult than traditional reference games, and we will show with a variety of metrics that the learned languages are more systematic, compositional, and interpretable. Finally, the rich compositional space of concepts explored in these games allows us to probe for specific logical operators in the emergent language. We propose a method for doing so, thereby demonstrating how the emergent languages reflect the compositional structure of their inputs.

\section{Communication Games}

First imagine a generic communication game between a teacher $T$ and student $S$. Let $G = (c, X^T, Y^T, X^S, Y^S)$ be a communication game, where $c : \mathcal{X} \mapsto \{0, 1\}$ is a latent concept to be communicated, $X^T = \{x^T_1, \dots, x^T_n\}$ is a set of $n$ inputs presented to the teacher, and $Y^T = \{y^T_1, \dots, y^T_n\}$ is a set of labels for the teachers' inputs, defined as $y^T_i = c(x^T_i)$. We call $x^T_i$ a \emph{target} if $y^T_i = 1$, which indicates that $x^T_i$ is a member of the concept $c$; otherwise $x^T_i$ is a \emph{distractor} and $y^T_i = 0$. $X^S$ and $Y^S$ are defined similarly for the student. Given its targets and distractors (but not the latent concept $c$), the teacher must send a message $m$ to a student that allows them to correctly identify their own targets, where $m = (m_1, \dots, m_n)$ is a discrete sequence over a fixed alphabet $m_i \in \mathcal{M}$. Now we can define variants of this communication game as follows: 

\paragraph{Reference game.} In basic reference games, the teacher and student see the same examples ($X^T = X^S$, $Y^T = Y^S$) and there is a single (repeated) target: $x^T_i = x^T_j$ for all $i, j$ where $y^T_i = y^T_j = 1$.\footnote{For the most consistent comparison across games, our reference game has multiple identical targets and student target decisions made independently for each input, instead of the single-target forced-choice setting. Appendix~\ref{app:refgame} shows results with traditional games trained with cross entropy loss; conclusions are the same.}

\paragraph{Set reference (setref) game.} Now we extend our game to \emph{sets}: the teacher and student see the same examples, but there are multiple target images encoding the concept (e.g.\ different \emph{red triangles}).

\paragraph{Concept game.} Finally, we propose the more abstract concept game, where the teacher and student see \emph{different examples} ($X^T \neq X^S$, $Y^T \neq Y^S$) of the same concept. When $X^T$ and $Y^T$ contain a single positive and negative example, this is a reference game with separate inputs for each agent, a setup which has been shown to encourage linguistic systematicity in some settings \cite{choi2018compositional,lazaridou2017multi,lazaridou2018emergence}.

\section{Models}
\label{sec:models}

Now we will formalize our models of the teacher and student. Given a communication game $G$, a teacher is defined as a distribution over messages given inputs $p^T(m \mid X^T, Y^T)$, and a student is a distribution over targets given a message: $p^S(Y^S \mid X^S, m) = \prod_i p^S(y^S_i \mid x^S_i, m)$. 

\paragraph{Teacher.} The teacher encodes all inputs with a convolutional neural network (CNN) $f^T_\theta$; embeddings for targets and distractors are averaged to form target and distractor \emph{prototype} embeddings \cite{snell2017prototypical},\footnote{Note that the teacher's job is one of representation learning for \emph{sets} \cite{zaheer2017deep} and thus one can use any set representation learning method beyond the prototypical networks explored here. As a more advanced implementation, we tried a variant of Set Transformers \cite{lee2019set}, but this did not give any tangible performance benefit.} which then conditions a recurrent neural network (RNN) used to produce the message. Let $X^T_+$ and $X^T_-$ denote the sets of targets and distractors in $X^T$; then define a prototype embedding $\mathbf{x}^T_+ = \frac{1}{|X^T_+|} \sum_{x_i \in X^T_+} f^T_\theta(x_i)$ (analogously for $\mathbf{x}^T_-$). Then $p^T(m \mid X^T, Y^T) = p_\textsc{RNN-Decode}(m \mid \proj([\mathbf{x}^T_+; \mathbf{x   }^T_-]))$ where $\proj$ is a linear projection to the RNN hidden state.

\paragraph{Student.} The student takes a message and makes predictions about the labels $\hat{y}^S_i$ independently for each input $x^S_i$. Given the teacher message and an input image, define $p^S(y^S_i \mid x^S_i, m) = \sigma(\textsc{RNN-Encode}(m) \cdot f^S_\phi(x^S_i))$, where $f^S_\phi$ is a separate CNN for the student.

We jointly train a teacher-student pair, including the vision modules and communication protocol, via stochastic gradient descent to maximize the student likelihood of selecting the correct target images. Formally, the loss for a single game is 
\begin{align}
    \mathcal{L}(T, S, G) = - \sum_i \log p^S(y^S_i \mid x^S_i, \hat{m}), \quad \hat{m} \sim p^T(m \mid X^T, Y^T).
    \label{eq:bce}
\end{align}
To maintain backwards differentiability, we use the straight-through Gumbel-Softmax \cite{Jang2017} trick with $\tau = 1$, simulating samples from the teacher distribution over tokens via softmax and discretizing in the forward pass only. For full model and training details and a link to code, see Appendix~\ref{app:modeldetails}.

\section{Tasks} 
\label{sec:tasks}

We examine the languages developed for our proposed communication games over two datasets: first, an artificial shape dataset which allows us to evaluate communication over cleanly defined logical concepts; second, a dataset of birds to test agents' ability to learn concepts from realistic visual input.

\begin{figure}[t]
    \centering
    \includegraphics[width=\linewidth]{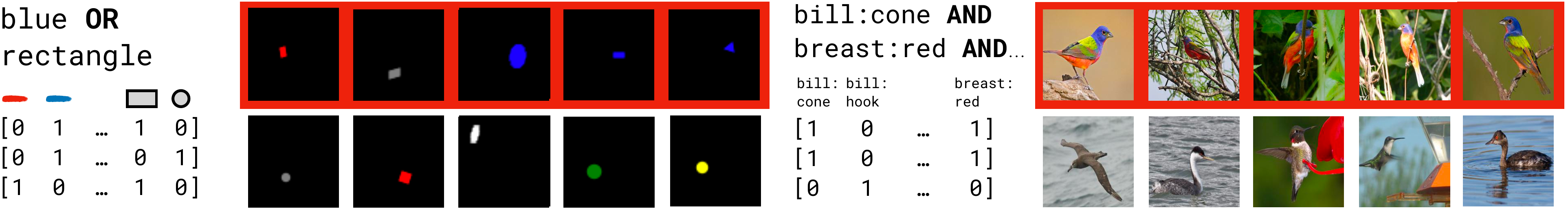}
    \caption{Example games, with targets (red border) and distractors for the ShapeWorld concept \emph{blue OR rectangle} (left) and the Birds concept \emph{painted bunting} (right). Concepts are represented as intensional logical formulas (top) or extensional sets of boolean input features (bottom), where each vector is the binary representation of an individual member of the concept (e.g.\ \emph{blue rectangle}, or the labeled attributes of one particular bird). See Figure~\ref{fig:detailed_datasets} in Appendix~\ref{app:modeldetails} for additional game examples.}
    \vspace{-1em}
    \label{fig:datasets}
\end{figure}

\paragraph{ShapeWorld.}
We use the ShapeWorld visual reasoning dataset \cite{Kuhnle2017} (Figure~\ref{fig:datasets}, left). For reference games, target images are a single object, defined by a conjunction of a shape and a color (e.g.\ \emph{red triangle}, \emph{green square}); of the 30 possible shapes, we hold out 20\% for testing.
For setref and concept games, concepts include the conjunctions tested in reference games, but also primitive concepts (e.g.\ \emph{blue shapes}) and arbitrary disjunctions or conjunctions of (possibly negated) shapes and/or colors.
This produces 312 concepts, 20\% of which are held out for testing.
These more general concepts cannot be tested in reference games, since a single object is always identified by a shape and color. This rules out disjunctive concepts like \emph{red OR blue} that only make sense across multiple objects. Similarly, since reference game targets must necessarily have both color \emph{and} shape, we can never guarantee that a message for a reference game only carries the semantics \emph{blue} and not, for example, \emph{blue circle}, if the target is a blue circle. By looking at sets, setref and concept games allow us to more precisely control the semantics of the concepts in each game.

Each game consists of 10 targets depicting shapes satisfying the concept, and 10 distractors. We specifically sample ``hard'' targets and distractors to test understanding of conjunctions or disjunctions (see Appendix~\ref{app:hard} for details).
Finally, we specify an agent vocabulary of 14 tokens and maximum length 5, so that the communication channel has the same bandwidth as the true concept formulas;\footnote{In the true concept formulas there are 5 shapes, 6 colors, and the 3 AND/OR/NOT operators, i.e.\ 14 tokens; and the longest concept formulas have the form \emph{NOT x AND NOT y}, i.e.\ length 5.} see Appendix~\ref{app:varylength} for experiments varying these parameters for both this dataset and the next one.

\paragraph{Birds.}
We next use the Caltech-UCSD Birds dataset \cite{wah2011caltech} which contains 200 classes of birds with 40--60 images (Figure~\ref{fig:datasets}, right). As before, reference games involve a single target; setref and concept game targets are members of a specific bird class. We use 100 classes at train and 50 at test, sampling 5 targets and 5 distractors per game.
The dataset contains boolean attributes (e.g.\ \emph{beak}, \emph{size}) for individual birds and classes.\footnote{Feature vectors for individual birds in a class vary due to the visibility of features in the image; class vectors are averaged across all individual birds, then rounded to 1 or 0.}  Thus, we represent reference game concepts as the feature vector of the target bird, and setref/concept game concepts as the feature vector of the class. In our evaluation, we will measure how well the languages capture these features. As there is no reference language for this task, we set the vocabulary size to 20 and the message length to 8 (though again see Appendix~\ref{app:varylength}).

\section{Evaluation}

We first measure communication success, as defined by student accuracy on held-out games from seen and unseen concepts, with the unseen concepts testing a language's ability to generalize compositionally. Ultimately, however, we are interested in the systematicity of the learned languages, which we evaluate via the following measures:

\paragraph{Information theoretic measures.} We first ignore the specific content of messages and concepts, and simply compute simple information theoretic quantities, by treating each distinct message and concept as a unique value and imagining probability distributions over these values. First, we measure the \textbf{conditional entropy} of teacher messages given concepts, $H(M \mid C)$, averaged across seen and unseen games; lower entropy indicates that agents use more consistent language for a fixed concept. However, $H(M \mid C)$ measures systematicity only in one direction; as a more symmetric measure, we also use the \textbf{adjusted mutual information}
\begin{align}
\AMI(M, C) = \left(I(M, C) - \mathbb{E}(I(M, C))\right) / \left(\max(H(M), H(C)) - \mathbb{E}(I(M, C))\right).
\end{align}
Here, $\mathbb{E}(I(M, C))$ is evaluated with respect to a hypergeometric model of randomness, where $M$ and $C$ are assumed to be random permutations subject to the number of unique values in either set \cite{vinh2010information}. AMI $\in [0, 1]$ represents the mutual information between $M$ and $C$, adjusted for chance to maintain comparability across different distributions of messages and concepts. A higher score indicates overall higher alignment between messages and concepts.

\paragraph{Topographic $\rho$.} To more precisely measure the lexical compositionality of a language, a measure often used in the literature is \emph{topographic $\rho$} \cite{brighton2006understanding,lazaridou2020emergent,lazaridou2018emergence,li2019ease},
which quantifies the agreement between two representational systems a la Representational Similarity Analysis \cite{kriegeskorte2008representational}.
We define a distance metric between game concepts $d_C(c_i, c_j)$ and another between agent messages $d_M(m_i, m_j)$, compute pairwise distances between concepts and between messages, and measure their alignment with Spearman's $\rho$. A high $\rho$ indicates that a teacher sends lexically similar messages for similar concepts.

For our distance function on messages, we use the \textbf{Edit} (i.e.~Levenshtein) distance with equal insert/delete/replace costs. For distances between game concepts, we define two distances based on intensional and extensional representations of concepts (Figure~\ref{fig:datasets}). First, we use the word-level \textbf{Edit} distance between string representations of logical formulas. Second, we use the \textbf{Hausdorff} distance $d_H$, a distance function between sets of members of a concept. Let $Z^a = \{ z^a_1, \dots, z^a_n \}$ be the set of feature-based boolean representations of inputs belonging to concept $a$. For ShapeWorld, these are two-hot vectors denoting the color and shape of all objects belonging to a specific formula; for example, for the concept \emph{red}, we have vectors for \emph{red triangle}, \emph{red circle}, and so on. For Birds, these are boolean vectors of the attributes of each individual bird of a species. Then the Hausdorff distance $d_H$ is the maximum distance from any point in one set to the closest point in the other:
$d_H(Z^a, Z^b) = \max(\sup_{i} d(z^a_i, Z^b), \sup_{j} d(z^b_j, Z^a))$,
where $d(a, B) = \inf_{b\in B} \, \EditDistance(a, b)$.

\section{Results}

\begin{table}[t]
\footnotesize
    \centering
    \caption{Student accuracy (seen and unseen concepts, where chance accuracy is 50\%), conditional entropy of messages given concepts (lower is better), and adjusted mutual information score (higher is better), with (SD) across 5 runs.
    }
    \vspace{1em}
    \begin{tabular}{llrrrr}
    \toprule
    Dataset & Game & Acc (Seen) & Acc (Unseen) & $H(M \mid C)$ & $\AMI(M, C)$ \\
         \midrule
    ShapeWorld & Ref & \textbf{97} (0.4) & \textbf{98} (0.3) & 7.3 (0.2) & 0.04 (0.00) \\
    & Setref & 92 (2.2) & 87 (1.6) & 3.9 (0.6) & 0.59 (0.08) \\
    & Concept & 88 (3.4) & 75 (3.0) & \textbf{2.4} (0.2) & \textbf{0.66} (0.07) \\
    \midrule
    Birds & Ref & \textbf{93} (0.3) & \textbf{89} (0.1) & 5.9 (0.2) & 0.05 (0.00) \\
    & Setref & 89 (0.2) & 78 (0.2) & 5.2 (0.1) & 0.17 (0.02) \\
    & Concept & 88 (0.1) & 73 (0.3) & \textbf{4.1} (0.2) & \textbf{0.26} (0.02) \\
         \bottomrule
    \end{tabular}
    \label{tab:perf}
    \vspace{-1em}
\end{table}

Table~\ref{tab:perf} shows test accuracy, as measured by student classification accuracy (partial credit given), on communication games over seen and unseen concepts for 5 models trained in each condition. Reference game performance is high across both datasets, and agents are able to generalize well to unseen games. Accuracy on setref and concept games is lower, with lower performance on novel games in both datasets.\footnote{To calibrate setref and concept performance, Appendix~\ref{app:upperbounds} tests listeners on ideal (human) languages.} Overall, communicating sets is a much harder task than specific reference.

\begin{figure}[t]
    \centering
    \includegraphics[width=\linewidth]{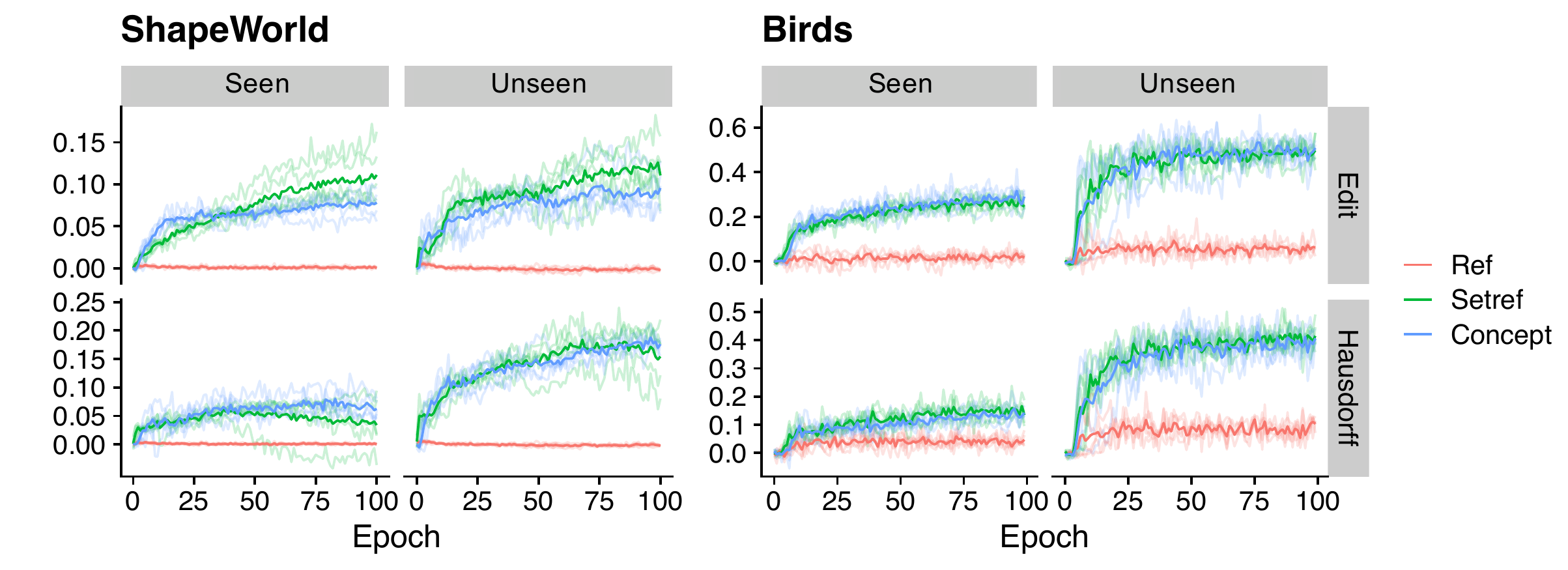}
    \caption{Topographic $\rho$ between language and Edit (top) or Hausdorff (bottom) concept distances for seen and unseen games across both datasets. Results from 5 runs plotted, with averages in bold.} 
    \label{fig:lang}
    \vspace{1em}
    \includegraphics[width=\linewidth]{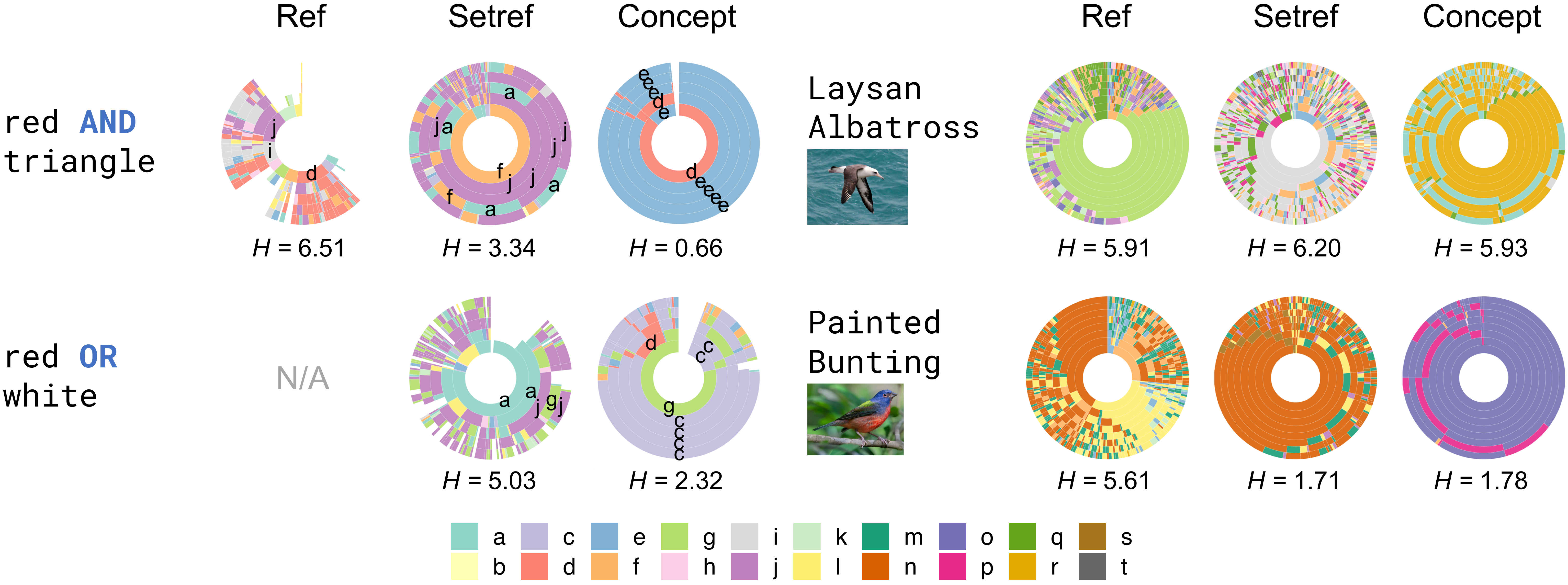}
    \caption{Distribution of 300 messages and entropies for game concepts in ref, setref, and concept settings. Messages start at the center and proceed outwards, with each colored section corresponding to a unique token and its frequency of occurrence at that position in the message. Empty spaces indicate end of sentence. For convenience, ShapeWorld plots are partially labeled with tokens.}
    \vspace{-1em}
    \label{fig:sunburst}
\end{figure}

The ability to communicate accurately, even on unseen concepts, is not necessarily indicative of more \emph{systematic} communication; generalization without compositional language is a common finding in the literature \cite{andreas2019measuring,chaabouni2020compositionality,kottur2017natural,lazaridou2018emergence}. Instead, we find that the more difficult games produce more systematic language.
For ShapeWorld, concept game entropy over messages is lower than reference game entropy (2.4 vs.\ 7.3), and $\AMI$ is higher (0.66 vs.\ 0.04), with setref in the middle; this pattern also occurs in Birds.
Furthermore, Figure~\ref{fig:lang} shows that topographic $\rho$ between the languages and the (Edit and Hausdorff) concept distances is higher for concept and setref than ref, throughout training.

Figure~\ref{fig:sunburst} (more examples in Appendix~\ref{app:sunburst}) shows messages generated by agents for concepts in both games, where we arbitrarily assign letters to agent ``words'' to assist with interpretation. For example, for concept game teachers conveying the concept \emph{red AND triangle}, the innermost circle is largely red, indicating that the majority of messages sent for this concept begin with \texttt{d}; proceeding outward, we see blue regions indicating the token \texttt{e}. Thus, concept agents consistently use the 5-token sequence \texttt{deeee} to refer to red triangles (less commonly, \texttt{edeee}). In contrast, for different red triangles, the language of reference game agents is strikingly inconsistent, as indicated by the extreme diversity of colors, while setref language is somewhere in the middle. The entropies over the message distributions for these concepts correlate with the ``busyness'' of the plots, reinforcing the idea that the setref and concept languages are more consistent.

\subsection{Set Size}
The difference between reference and setref/concept games can be interpreted as a continuum, ranging from referring to a single object (reference) to referring to a potentially infinite number of objects. With a small number of objects, it may still be possible to communicate only low-level, non-generalizable features of the set, similar to the strategies adopted by our reference game agents. In contrast, increasingly large numbers of objects should put further pressures on the semantics of the messages to not refer to individual inputs, but rather entire categories.
\begin{wrapfigure}{r}{0.5\textwidth}
    \centering
    \includegraphics[width=\linewidth]{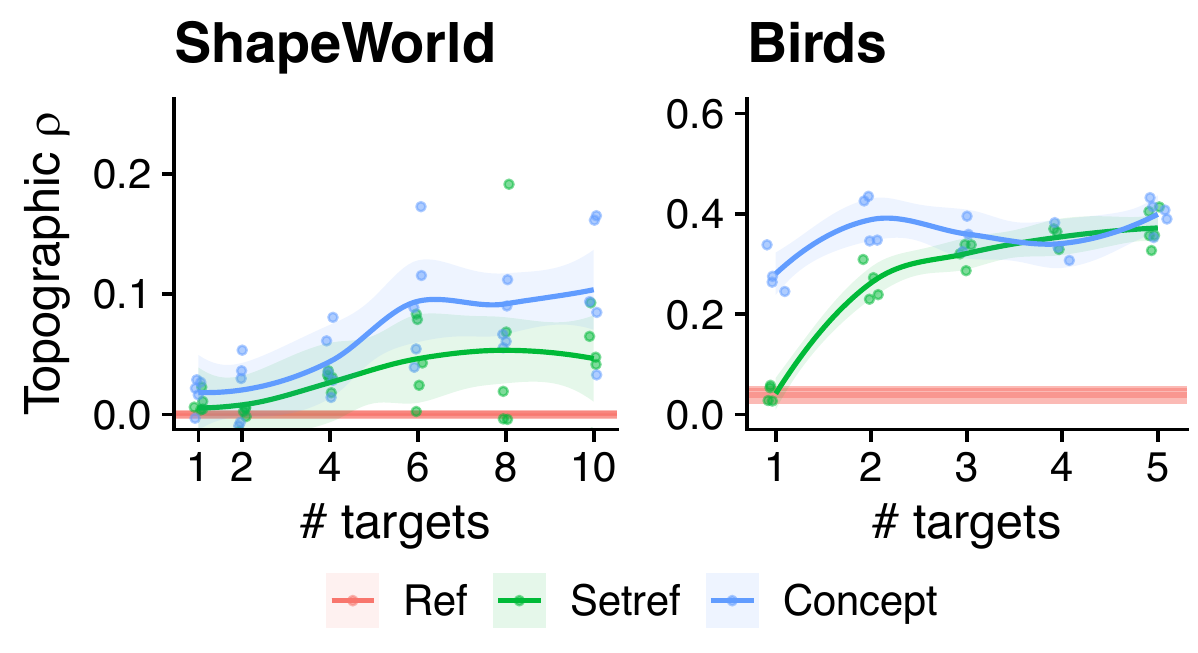}
    \caption{Topographic $\rho$ (concept Edit distance) with varying number of targets (average over seen/unseen splits). Each point and Ref line is an independent run.}
    \label{fig:n_ex}
    \vspace{-1em}
\end{wrapfigure}

In Figure~\ref{fig:n_ex}, we confirm this hypothesis, showing how increasing the number of targets  $n$ (with equal numbers of distractors) increases language systematicity. $n$ has a statistically significant effect on topographic $\rho$ for ShapeWorld setref (Spearman $\rho = 0.39, p = 0.029$) and concept ($\rho = 0.75, p < 10^{-5}$) and Birds setref ($\rho = 0.90, p < 10^{-7}$) and concept ($\rho = 0.51, p = 0.024$). When $n = 1$, the setref game is equivalent to a reference game with 1 target and 1 distractor, and the concept game is similar, but with agents given separate inputs. Our results suggest that this decoupling, as proposed by \citet{lazaridou2017multi} and often used in the emergent communication literature, promotes systematicity in some cases (Birds) \emph{but not others} (ShapeWorld). We additionally show that (1) sets are an \emph{alternative} way of encouraging systematicity without needing this separation, and (2) even with this separation, larger set sizes further improve the systematicity of the resulting language.

\subsection{Generalizing across game types}
To measure the generality of the agents' strategies, we evaluate their ability to generalize zero-shot across game types. Table~\ref{tab:xeval} shows accuracy and systematicity metrics for agents evaluated on games of a different type. Agents trained on the setref and concept games are able to generalize to reference games (\textcolor{darkyellow}{yellow cells}), producing systematic referring expressions, as they have already been biased towards generic language. Importantly, setref game agents can generalize to concept games with separate inputs (\textcolor{darkmagenta}{magenta cells}), though to a lesser extent for Birds (75\% vs 84\%). This suggests that sets pressure agents to learn generalizable features \emph{despite} the shared input. In contrast, we see little generalization ability from reference games to setref and concept games (\textcolor{darkorange}{orange cells}), suggesting that agents are not conveying generalizable features, but rather spurious patterns in the input \cite{bouchacourt2018agents}.

\begin{table}[t]
    \centering
    \caption{\triple{Accuracy}{AMI}{Topographic $\rho$ (concept Edit distance)} for agents \emph{trained} on different game types (columns), then \emph{evaluated} (zero-shot) on different game types (rows). Chance accuracy is 50\%. Gray shaded cells indicate standard test-time evaluation; other cell colors are explained in the text. Note that for ShapeWorld reference agents, we evaluate only on setref and concept games that use the 30 conjunctive concepts tested in reference games (e.g.\ \emph{red triangle}, \emph{blue square}).}
    \vspace{1em}
\footnotesize
    \begin{tabular}{lrrr}
    \toprule
     & Train Ref & Train Setref & Train Concept \\
         \midrule
        \textbf{ShapeWorld} & & & \\
         Eval Ref & \graycell \triple{98 (0.4)}{.04 (.00)}{.00 (.00)}  & \yellowcell \triple{91 (4.3)}{.43 (.09)}{.42 (.11)}  & \yellowcell \triple{83 (5.6)}{.60 (.04)}{.63 (.08)} \\
         Eval Setref & \orangecell \triple{56 (0.1)}{.02 (.00)}{.00 (.00)} & \graycell \triple{90 (1.3)}{.59 (.08)}{.12 (.03)} & \triple{83 (6.7)}{.66 (.07)}{.09 (.01)} \\
         Eval Concept & \orangecell \triple{50 (0.0)}{.02 (.00)}{.00 (.00)}  & \magentacell \triple{90 (4.6)}{.59 (.08)}{.12 (.03)} & \graycell \triple{82 (3.0)}{.66 (.07)}{.09 (.01)} \\
         \midrule
        \textbf{Birds} & & & \\
                 Eval Ref & \graycell \triple{91 (0.8)}{.05 (.00)}{.04 (.01)}  & \yellowcell \triple{85 (1.1)}{.14 (.02)}{.16 (.02)}  & \yellowcell \triple{82 (0.9)}{.20 (.02)}{.15 (.01)} \\
         Eval Setref & \orangecell \triple{64 (1.7)}{.03 (.01)}{.15 (.01)} & \graycell \triple{84 (1.1)}{.17 (.02)}{.37 (.04)} & \triple{78 (0.7)}{.26 (.02)}{.40 (.03)} \\
         Eval Concept & \orangecell \triple{56 (0.9)}{.03 (.01)}{.15 (.01)}  & \magentacell \triple{75 (0.9)}{.17 (.02)}{.37 (.04)} & \graycell \triple{82 (0.6)}{.26 (.02)}{.40 (.03)} \\
         \bottomrule
    \end{tabular}
    \label{tab:xeval}
    \vspace{-1em}
\end{table}

\section{Probing for compositionality}
\label{sec:acre}

The richer set of concepts afforded by our setref and concept games allow for more detailed analyses of the emergent languages beyond the standard metrics presented above.
Broadly, we are interested in whether the compositional structure of concepts is reflected in the language. For example, our agents produce messages for the primitive concepts \emph{red} and \emph{triangle}, as well as the conjunctive concept \emph{red AND triangle}.\footnote{We cannot do this analysis for reference games, since we cannot test primitive concepts; recall Section~\ref{sec:tasks}.} Natural language is equipped with a composition operator, $\AND(m_1, m_2) = m_1 \; \texttt{AND} \; m_2$, that operates solely on lexical forms
and whose meaning is defined as the conjunction of the meanings of its arguments. Does a similar operator exist in the emergent language (Figure~\ref{fig:acre_overview}a)?

We propose to \emph{learn} such an operator by training a model to compose messages in the emergent language to form their conjunctions. Like the English $\AND$, this operator must be able to combine any of the concepts in the language, and must crucially \emph{generalize} to novel combinations of features. If, given new concepts, we can reconstruct a message that induces the right behavior in the student, this suggests our model has learned some analog of $\AND$ in the language. The reconstruction accuracy of this model can then be interpreted as a much more explicit measure of compositionality than the measures explored above: it reveals the degree to which the syntax of the emergent language operationalizes the specific logical operations present in the underlying space of concepts.

Our method is inspired by the \emph{Tree Reconstruction Error} (TRE)
metric proposed by Andreas \cite{andreas2019measuring}, which learns a compositional approximation to a representation space, assuming the latent compositional structure is known.\footnote{We cannot apply TRE directly to our setting. Andreas \cite{andreas2019measuring} applied TRE to a standard reference game, where targets are represented as conjunctions of shape and color features (e.g.\ \emph{blue square}, \emph{red triangle}). As we mention in Section~\ref{sec:tasks}, because reference games cannot test primitive concepts like \emph{red} and \emph{triangle}, Andreas \cite{andreas2019measuring} proposed to \emph{learn} representations for primitives which can then be composed via some (predefined) composition function. However, in our setting, it makes little sense to learn arbitrary representations for primitive concepts, when we actually have real messages for such concepts in the first place, hence the method we propose.} However, there are several crucial differences in our method that are optimized for probing for linguistic structure. First, we \emph{learn} a set of arbitrary composition operations, instead of imposing a predefined operation (e.g.\ elementwise addition). Moreover, these learned composition operators are actually valid and interpretable linguistic transformations on messages, rather than operations (like addition) that work only on internal representations of sub-concepts. And finally, we evaluate our operators on held-out concepts, examining how the reconstructed messages serve the ultimate communicative goal: inducing the correct generalization behavior in the student.

\subsection{Learning an Approximate Compositional Reconstruction (ACRe)}

Let us first formalize the learning problem: after training our agents, we have a dataset of message and concept pairs $\mathcal{T} = \{ (m_i, c_i) \}$ generated by a teacher for each game. Each concept is one of a set of logical forms $\mathcal{L}(\mathcal{C})$ defined inductively over a set of \emph{primitive} concepts $\mathcal{C}$ (e.g.\ \emph{red}, \emph{triangle}) and \emph{composition operations} $\Omega$ as follows:
\begin{enumerate}
    \item Every primitive concept is in $\mathcal{L}(\mathcal{C})$: $\mathcal{C} \subseteq \mathcal{L}(\mathcal{C})$.
    \item Every composition of concepts is a concept: let $\Omega_n$ be the set of $n$-ary composition operations. Then $\forall n,\; (c_1, c_2, \dots, c_n) \in \mathcal{L}(\mathcal{C})^n,\; \omega \in \Omega_n$, we have $\omega(c_1, c_2, \dots, c_n) \in \mathcal{L}({\mathcal{C}})$.
\end{enumerate}

$\mathcal{T}$ defines a probability distribution over messages given concepts: $p_\mathcal{T}(m \mid c)$.
Our aim is to learn an \textbf{A}pproximate \textbf{C}ompositional \textbf{Re}construction to these messages $\hat{p}_\eta(m \mid c)$, composed of $\eta$-parameterized message distributions that factorize along the compositional structure of $\mathcal{L}({C})$:
\begin{align}
\hat{p}_\eta(m \mid c) = \begin{cases}
    \hat{p}^c_\eta(m) & \text{if $c \in \mathcal{C}$, i.e.\ $c$ is primitive} \\
    \mathbb{E}_{\hat{m}_i \sim \hat{p}_\eta(m_i \mid c_i)} \left[ \hat{p}^\omega_\eta (m \mid \hat{m}_1, \dots, \hat{m}_n)  \right] & \text{if $c = \omega(c_1, \dots, c_n)$}, 
\end{cases}
\label{eq:acre}
\end{align}

where $\hat{p}^c_\eta(m)$ is a model of the distribution of messages for primitive concept $c$, and $\hat{p}^\omega_\eta(m \mid m_1, \dots, m_n)$ is a learned lexical analog to operation $\omega$ that takes in messages and outputs a distribution over composed messages. Given a concept $c$, we can sample a message from $\hat{p}_\eta(m \mid c)$ by either sampling directly from $\hat{p}^c_\eta$ (if $c$ is primitive) or recursively sampling messages $m_i$ from the constituents of $c$, then sampling from the corresponding $\hat{p}^\omega_\eta$.

We implement $\hat{p}^c_\eta$ and $\hat{p}^\omega_\eta$ as small 2-layer transformer-based language models (LMs) \cite{vaswani2017attention}. For each $c$, $\hat{p}^c_\eta$ is an unconditional LM (i.e.\ we have an LM for \emph{red}, \emph{triangle}, etc.). For $n$-ary operations $\omega$, $\hat{p}^\omega_\eta$ is a sequence-to-sequence (seq2seq) model decoding from the concatenated arguments: $\hat{p}^\omega_\eta(m \mid m_1, \dots, m_n) = p^{\text{decode}}_{\eta}(m \mid m_1 \; \texttt{[SEP]} \; m_2 \; \dots \; \texttt{[SEP]} \; m_n)$.
Note that this formulation imposes few constraints on the mechanism of the composition operator: for example, we are not enforcing that there exists a token denoting conjunction (\texttt{AND}), or that the arguments must be copied verbatim into the output. These constraints, generally true of human languages, could be explored in future work.

To train ACRe, let $\mathcal{T}_c = \{ (m_i, c_i) \in \mathcal{T} \mid c_i = c\}$ be the set of message-concept pairs with concept $c$, and $\mathcal{T}_\omega = \{ (m_i, c_i) \in \mathcal{T} \mid c_i = \omega(\cdot) \}$ be the set of messages where $c_i$ uses $\omega$.
Now, for each primitive $c \in \mathcal{C}$, train the LM $\hat{p}^c_\eta$ on $\mathcal{T}_c$ to approximate $p_\mathcal{T}(m \mid c)$. Then, freezing these models, for $n = 1, \dots, N$, $\omega \in \Omega_n$, train the composition model $\hat{p}^\omega_\eta$ on $\mathcal{T}_\omega$. Specifically, given a pair $\left(m, \omega(c_1, \dots, c_n)\right)$, first sample messages for the sub-concepts $c_i$ from our frozen models: $\hat{m_i} \sim \hat{p}_\eta(m_i \mid c_i)$.\footnote{As presented, this procedure is only possible if the composition models can be trained and frozen in a specific order without circular dependencies. For example, we cannot naively train $\hat{p}^\omega_\eta$ on concepts of the form $\omega(\omega(\cdot))$, since sampling from the inner $\omega$ requires an already-trained $\hat{p}^\omega_\eta$. Learning from arbitrary concepts is possible by backpropagating through samples (e.g. via Gumbel-Softmax), which we leave for future work.} Then, train the composition model to maximize $\hat{p}^{\omega}_\eta(m \mid \hat{m_1}, \dots, \hat{m}_n)$. For example, given a message \texttt{bca} for the concept $\emph{red AND triangle}$, we first sample messages for \emph{red} and \emph{triangle} from $\hat{p}^\text{red}_\eta$ and $\hat{p}^\text{triangle}_\eta$, then train $\hat{p}^\text{AND}_\eta$ to decode \texttt{bca} from these messages (Figure~\ref{fig:acre_overview}). Full training and model details are in Appendix~\ref{app:acredetails}.

\begin{figure}[t]
    \centering
    \includegraphics[width=\textwidth]{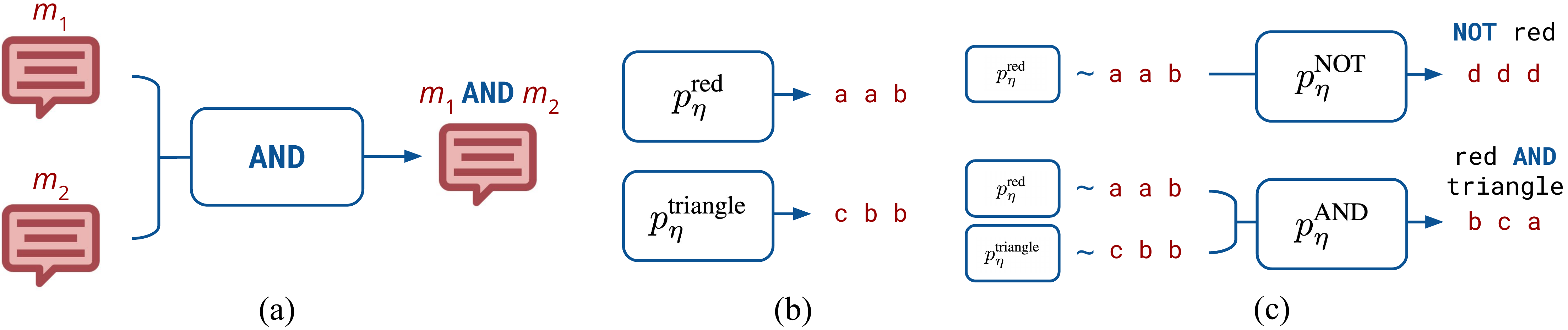}
    \caption{Our ACRe procedure. (a) Does a lexical analog of AND exist in our emergent language? (b) We first train primitive LMs to mimic the distribution of agent messages given a fixed concept. (c) We then train composition operations by sampling arguments from primitive LMs, then training a seq2seq model to mimic the agent message produced for a higher-order concept.}
    \label{fig:acre_overview}
    \vspace{-1em}
\end{figure}

\subsection{Evaluation and results}

After training, each $\hat{p}^\omega_\eta$ is a learned lexical analog to the composition operation $\omega$. To test whether our approximation has learned the semantics of the language, we hold out 20\% of conjunctive and disjunctive concepts $c'_i$ during training and sample messages for these concepts from the learned $\hat{p}_\eta(m \mid c'_i)$ according to Equation~\ref{eq:acre}. We evaluate how well these reconstructed messages encode the concepts via (1) lexical overlap with the true teacher messages (as measured by BLEU) and (2) student performance when given our language and a game for the corresponding concept.\footnote{In these experiments, we run ACRe on agents trained on the full set of 312 concepts in ShapeWorld, since we are not testing compositional generalization of the agents, but of ACRe.}

\begin{table}[t]
\footnotesize
    \centering
    \caption{ACRe evaluation. (SD) across 5 runs. In the highlighted cells we conduct paired $t$-tests, comparing ACRe to Closest; * indicates significance at $p < 0.05$.}
    \vspace{1em}
    \begin{tabular}{llrrrrrr}
    \toprule
    & & \multicolumn{3}{c}{\textbf{Train}} & \multicolumn{3}{c}{\textbf{Test}} \\
    Game & Language & BLEU-1 & BLEU-4 & Student Acc & BLEU-1 & BLEU-4 & Student Acc \\
         \midrule
    \textbf{Setref} & Teacher & 100 (0.0) & 100 (0.0) & 91 (2.7)  & 100 (0.0)\hphantom{*} & 100 (0.0)\hphantom{*} & 86 (5.2)\hphantom{*}\\
      & ACRe & 96 (2.1) & 73 (5.0) & 81 (3.4) & \graycell 91 (3.9)* & \graycell 52 (10.0)* & \graycell 65 (3.4)* \\
      & Closest & 78 (4.6) & 28 (6.2) & 48 (0.7) & 88 (3.8)\hphantom{*} & 38 (9.0)\hphantom{*} & 56 (0.7)\hphantom{*}\\
      & Random & 71 (6.6) & 22 (4.3) & 50 (0.0) & 73 (7.4)\hphantom{*} & 24 (5.5)\hphantom{*} & 50 (0.3)\hphantom{*}\\
    \midrule
    \textbf{Concept} & Teacher & 100 (0.0) & 100 (0.0) & 88 (4.2) & 100 (0.0)\hphantom{*} & 100 (0.0)\hphantom{*} & 84 (4.4)\hphantom{*} \\
      & ACRe & 95 (3.4) & 80 (7.6) & 82 (2.2) & \graycell 87 (6.9)* & \graycell 59 (12.2)* & \graycell 70 (4.0)* \\
      & Closest & 64 (3.7) & 28 (6.2) & 48 (1.6) & 76 (3.2)\hphantom{*} & 38 (6.7)\hphantom{*} & 56 (1.3)\hphantom{*}\\
      & Random & 54 (3.0) & 19 (2.8) & 50 (0.2) & 56 (2.2)\hphantom{*} & 20 (3.2)\hphantom{*} & 50 (0.4)\hphantom{*} \\
    \bottomrule
    \end{tabular}
    \label{tab:acre}
\end{table}

Results are in Table~\ref{tab:acre}. The upper bound on performance is given by the true \textbf{Teacher} language, $p_\mathcal{T}(m \mid c)$. We also use language sampled randomly from (1) teacher messages for \emph{any} concept $p_\mathcal{T}(m)$ (\textbf{Random}) and (2) teacher messages for the \textbf{Closest} concept as measured by Edit distance (breaking ties randomly). ACRe's performance is a measure of the degree of compositionality in the language, upper bounded by the teacher language and lower bounded by the random baseline. The results reveal some degree of compositional structure: ACRe reconstructs the training data well and crucially outperforms baselines in predicting messages for unseen concepts. Of course, our ACRe model trails Teacher language accuracy by 16--20 points. This gap could stem from a failure of ACRe to find the correct compositional generalization, or lack of compositionality in the language itself.

\begin{figure}[t]
    \centering
    \includegraphics[width=0.8\textwidth]{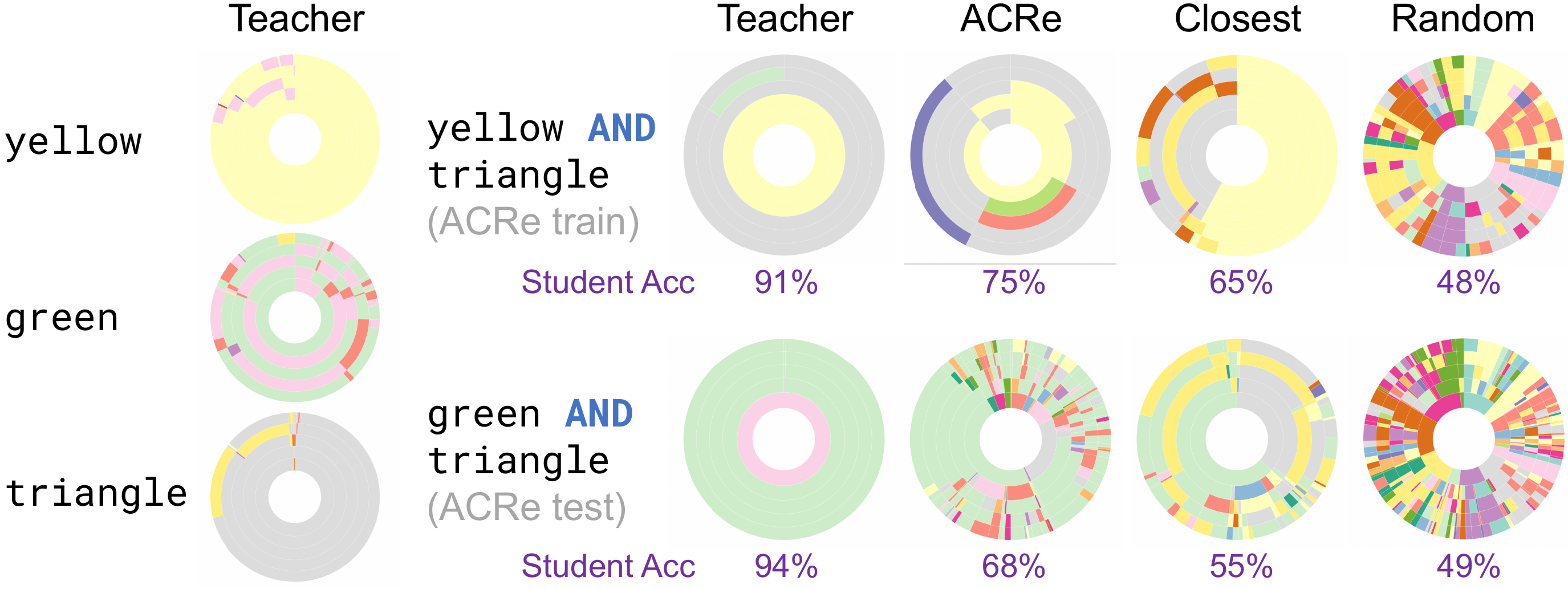}
    \caption{Composition in the emergent languages. A concept game teacher's messages for primitive concepts \emph{yellow}, \emph{green}, and \emph{triangle}, conjunctions \emph{yellow AND triangle} and \emph{green AND triangle} (where \emph{green AND triangle} is an unseen combination for ACRe), and predicted messages according to ACRe and other baselines. Color key same as Figure~\ref{fig:sunburst}.}
    \label{fig:acre_example}
    \vspace{-1em}
\end{figure}

A qualitative analysis supports the interpretation that the compositional operations are not always interpretable. Figure~\ref{fig:acre_example} shows an example of message distributions for the primitive concepts \emph{yellow}, \emph{green}, and \emph{triangle}, as well as distributions for the conjunctive concepts \emph{yellow AND triangle} and \emph{green AND triangle}, with predictions from ACRe and baseline models. The message for \emph{yellow AND triangle} is intuitively composed out of tokens concatenated from both primitives similar to natural language \cite{haiman1980iconicity}.
However, the message \emph{green AND triangle} uses tokens from \emph{green}, but none from \emph{triangle}, thereby violating the \emph{mutual exclusivity} principle of language \cite{markman1988children}. Regardless, in both cases our ACRe model is able to approximate such messages better than the other baselines. An exciting avenue for future work is encouraging models to develop operators more akin to human language, and evaluating their acquisition by searching among a more restricted class of ACRe models $\hat{p}^\omega_\eta$.

\section{Related Work}

\paragraph{Promoting compositionality in multi-agent communication.} Compositionality and systematicity in emergent languages have long been \emph{a priori} goals in multi-agent communication. Such languages may be easier to interpret, teach, and integrate with human language via supervised learning \cite{lowe2020interaction}. Towards this end, a large body of work (see \cite{lazaridou2020emergent} for review) has explored what factors might encourage compositionality in emergent communication, such as teachability and iterated learning \cite{li2019ease,ren2020compositional}, agent capacity \cite{resnick2020capacity} and regularization \cite{luna2020internal}, self-understanding \cite{choi2018compositional}, and game design \cite{kottur2017natural,lazaridou2018emergence}. However, most of this existing work operates within the limited paradigm of the Lewis \cite{Lewis1969} reference game. In this paper, we propose to revisit and revise the fundamentally limited reference game objective: inspired by human teaching \cite{chopra2019ratchet} and generic language \cite{tessler2019language}, we encourage our agents to communicate \emph{generalizations} over objects, which significantly increases linguistic systematicity, orthogonal to any of the alternative pressures proposed in related work.

\paragraph{Measures of compositionality in languages.} Crucial to the problem of promoting compositionality in emergent communication is how to measure it in the first place. The literature has seen a wide variety of methods \cite{andreas2019measuring,andreas2017analogs,brighton2006understanding,chaabouni2020compositionality,resnick2020capacity} claiming to more accurately align with human notions of compositionality, some of which are reported here. Most of this existing work focuses on outputting a broad scalar quantity that represents the degree of compositionality in a language \cite[e.g.\ topographic $\rho$;][]{brighton2006understanding}. In contrast, ACRe is a much more granular attempt at measuring not just \emph{how much} compositionality, but \emph{what kinds} of compositionality emerge, by actually learning, evaluating, and interpreting each distinct compositional operation.

This brings our work more in line with more precise studies of the emergence of composition operations in emergent languages \cite{steinert2016compositional,steinert2020toward} and the analyses of van der Wal \emph{et al.} \cite{van2020grammar} and Andreas \cite{andreas2019measuring}. In contrast to Steinert-Threlkeld \cite{steinert2016compositional,steinert2020toward}, who studies simpler settings where compositionality can be verified with manual inspection, we propose a way to measure compositionality in more complex languages that clearly do not exhibit perfect compositionality, but may still have learnable latent structure. In contrast to van der Wal \emph{et al.} \cite{van2020grammar}, who use grammar induction techniques for syntactic analysis of emergent languages, we tailor our syntactic analysis to messages generated for a known semantic space of concepts. This lets us approximate concrete syntactic operations in the language, and evaluate how well the approximations capture the corresponding semantic operations. Lastly, our method ACRe builds off of the TRE metric developed by Andreas \cite{andreas2019measuring}, and we describe this relationship in Section~\ref{sec:acre}.

\paragraph{ACRe as program induction and grammatical inference.} Finally, ACRe is reminiscent of a program induction or grammatical inference problem, where inputs are agent messages for primitive concepts, and outputs are the messages produced after some composition has been applied to the inputs. Our task is to discover the (ideally simple and interpretable) lexical programs that implement the corresponding transformation in semantic space. Because we have no priors over what lexical transformations, if any, the emergent languages might implement, we search for programs among a general class of seq2seq translation models. However, in this domain, human languages have much simpler lexical operators involving concatenation and infix notiation (e.g. \texttt{x AND y}), and in the future, we would like to push emergent languages towards stricter compositionality. One way of benchmarking more cleanly compositional languages is to restrict ACRe models to more constrained and interpretable programs learned with techniques from the program synthesis \cite{gulwani2011automating} or grammar induction \cite{cicchello2003inducing} literature.

\section{Conclusion}
\label{sec:conclusion}

We have proposed extensions of referential games to sets of objects, and found that the need to convey generalizable categories leads to the development of more systematic languages, whether inputs are shared (setref) or unshared (concept) across agents.
Moving forward, the richer space of concepts afforded by our setref and concept games are a promising testbed for studying the emergence of higher-level linguistic phenomena, such as quantifiers or probabilistic language. Finally, while ACRe reveals some compositional structure in the emergent languages as-is, the learned composition operations are not particularly interpretable. Future work should identify what kinds of environmental or architectural constraints might encourage more transparent composition in learned artificial languages. One challenging evaluation of compositionality along these lines is to measure the ability of agents to extrapolate to longer and more complex concepts not seen during training (e.g.\ \emph{green OR (blue AND triangle)}), and evaluating ACRe's ability to capture this recursive structure.

\section{Broader Impact}
\label{sec:broaderimpacts}

Our work investigates agents communicating in artificial and isolated environments, and thus has limited immediate societal impact. However, we can imagine that with future advances in research and compute, agents may learn linguistic strategies to collaborate on real-world problems in increasingly high-stakes domains, and it will be important to ensure that the learned languages are interpretable, safe, and reliable. Our work has potential positive benefits in these scenarios: our proposed games encourage agents to learn more human-like linguistic behavior, which might ease collaboration with humans; and ACRe is a tool for evaluating and interpreting learned languages. However, future work is needed to see whether these tools remain reliable as we scale up our agents and tasks.

\begin{ack}
We thank Alex Tamkin, Elisa Kreiss, Josh Rozner, Gabriel Poesia, and Chris Potts for helpful comments and discussions, and our anonymous reviewers for extensive feedback on the paper. This research was supported by an NSF Graduate Research Fellowship for JM, the Stanford HAI-AWS Cloud Credits for Research program, and the Office of Naval Research grant ONR MURI N00014-16-1-2007.
\end{ack}

\small 

\bibliography{emergent_generalization_neurips21.bib}
\bibliographystyle{abbrvnat}

\newpage
\normalsize
\appendix

\renewcommand\thefigure{S\arabic{figure}}
\renewcommand\thetable{S\arabic{table}}
\setcounter{figure}{0}
\setcounter{table}{0}

\section{Model, training, and dataset details}
\label{app:modeldetails}

All models are trained end-to-end with the Gumbel-Softmax \cite{Jang2017} trick with the Adam \cite{kingma2014adam} optimizer with learning rate 0.0001. Models are trained on a single Titan Xp GPU on an internal cluster. Training time is typically 6-8 hours on 4 CPUs and 32GB of RAM.
Code and data are available at \url{https://github.com/jayelm/emergent-generalization}.

\subsection{ShapeWorld}

\paragraph{Model.} $f^T_\theta$ and $f^S_\phi$ are 4-layer convolutional neural networks, each consisting of a 64-filter 3x3 convolution, batch normalization, ReLU nonlinearity, and 2x2 max-pooling layer, as used in the few-shot learning literature \cite{snell2017prototypical}. RNN encoders and decoders are single layer Gated Recurrent Units (GRUs) \cite{cho2014learning} with hidden size 1024 and embedding size 500. We train with batch size $B = 128$.

We noticed that for ShapeWorld specifically, our setref and concept games easily converged to local minima with approximately 83\% maximum accuracy by only considering color features and ignoring shape features. In these experiments, we had speakers sample tokens from a mixture of 90\% the original logits, and 10\% a uniform distribution over tokens, to encourage exploration (similar to $\varepsilon$-greedy policies in RL), which improved performance across all games.

\paragraph{Data.} As aforementioned, for reference games there are 30 concepts (conjunctions of shape and color e.g.\ \emph{blue square}, \emph{green ellipse}), and for setref and concept games there are 312 total concepts (see Appendix~\ref{app:hard}), of which 80\% are reserved for training and 20\% are reserved for test. From the training concepts, we sample 20,000 base games to use as our training set, each with 40 targets and distractors each. At training time, we perform augmentation by randomly selecting 10 targets and 10 distractors given to both teacher and student, meaning that the total set of games is combinatorially large (over $\binom{50}{10}$ combinations for reference games, and $\binom{50}{10}^2$ combinations for setref and concept). We set up validation and test datasets with 2000 games each, divided among seen and unseen concepts, with no augmentation performed. We train over 100 epochs (defined by a single pass through 20,000 augmented games) until average performance on the validation set is maximized.

\paragraph{License.} ShapeWorld is distributed under an MIT license (\url{https://github.com/AlexKuhnle/ShapeWorld/blob/master/LICENSE}).

\subsection{Birds}

\paragraph{Model.} $f^T_\theta$ and $f^S_\phi$ is an ImageNet \cite{russakovsky2015imagenet}-pretrained ResNet-18 \cite{he2016deep}; similar results were observed for models trained from scratch. Like ShapeWorld, RNN encoders and decoders are single layer GRUs with hidden size 1024 and embedding size 500. We train with batch size $B = 16$ and preprocess images with ImageNet mean normalization.

\paragraph{Data.} From the 100 training classes, we sample games dynamically by randomly selecting 5 positive targets from the class and 5 negative targets randomly. Like in ShapeWorld, this makes the number of possible training games combinatorially large. We set up validation and test datasets with 400 games each divided among seen and unseen concepts. We define an epoch as a single pass through 1,000 augmented training games, and like before, select the model with the highest performance on the validation set after 100 epochs.

\begin{figure*}[ht]
    \centering
    \includegraphics[width=\textwidth]{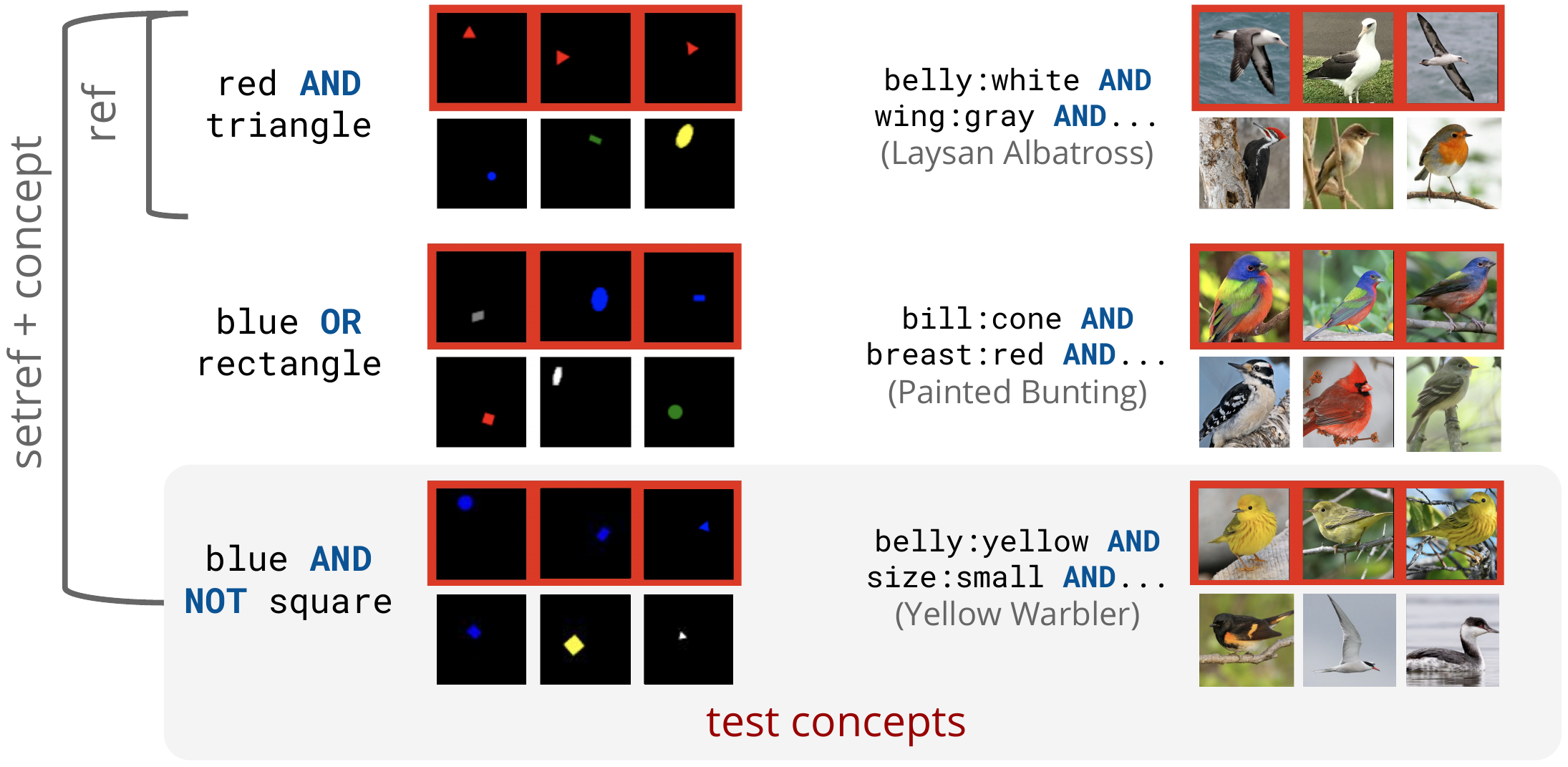}
    \caption{More example games for both ShapeWorld and Birds datasets.} 
    \label{fig:detailed_datasets}
    \vspace{-1em}
\end{figure*}

For additional example games from both datasets, see Figure~\ref{fig:detailed_datasets}.

\paragraph{License.} Caltech-UCSD Birds dataset is not distributed under a license; images are retrieved from Flickr and are the property of the respective photographers (\url{http://www.vision.caltech.edu/visipedia/CUB-200-2011.html}).

\section{ShapeWorld concepts}
\label{app:hard}

The 312 ShapeWorld concepts are either:

\begin{enumerate}
    \item A single \emph{primitive} shape (\emph{triangle, square, circle, ellipse, rectangle}) or color (\emph{red, blue, green, yellow, white, gray}), possibly negated (e.g.\ \emph{not gray});
    \item A disjunction of two (possibly negated) primitives (e.g.\ \emph{blue or yellow}, \emph{circle or not red});
    \item A conjunction of two (possibly negated) primitives (e.g.\ \emph{red and triangle}, \emph{red and not triangle}).
\end{enumerate}

We enumerate all (boolean-equivalent) possible formulas, then discard formulas which are tautologically true (e.g.\ \emph{not yellow or not red}) or unsatisfiable (e.g.\ \emph{circle and square}).

For each concept, sampling positive and negative shapes uniformly often results in games that do not specifically test the concept. For example, for the concept \emph{gray and not circle}, there may not be any negative \emph{gray circles}, so the agent could just infer the concept \emph{gray}. To ensure that concepts are fully tested, for disjunctive concepts, we sample 1/3 targets that satisfy \emph{only} the \emph{left} side of the disjunction; 1/3 that satisfy only the \emph{right} side; and 1/3 that satisfy both. For conjunctions, we sample 1/3 \emph{distractors} that only fail to satisfy the left side of the disjunction; 1/3 that only fail to satisfy the right side; and 1/3 that fail to satisfy both sides.

Code used for generating the dataset is available at \url{https://github.com/jayelm/minishapeworld/tree/neurips2021}.

\section{Varying communication channel size}
\label{app:varylength}

To see whether our results hold as we vary the bandwidth of the communication channel, we run agents on both datasets with the following configurations of (vocabulary size, max message length):

\begin{itemize}
    \item \textbf{Small} (S): (3, 3), leading to only 27 possible messages, which is not enough messages to uniquely identify each concept in any of the games explored here
    \item \textbf{Medium} (M): (5, 5)
    \item \textbf{Large} (L): (100, 20)
    \item \textbf{X-Large} (XL): (1000, 20)
\end{itemize}

\begin{figure}
    \centering
    \includegraphics[width=\textwidth]{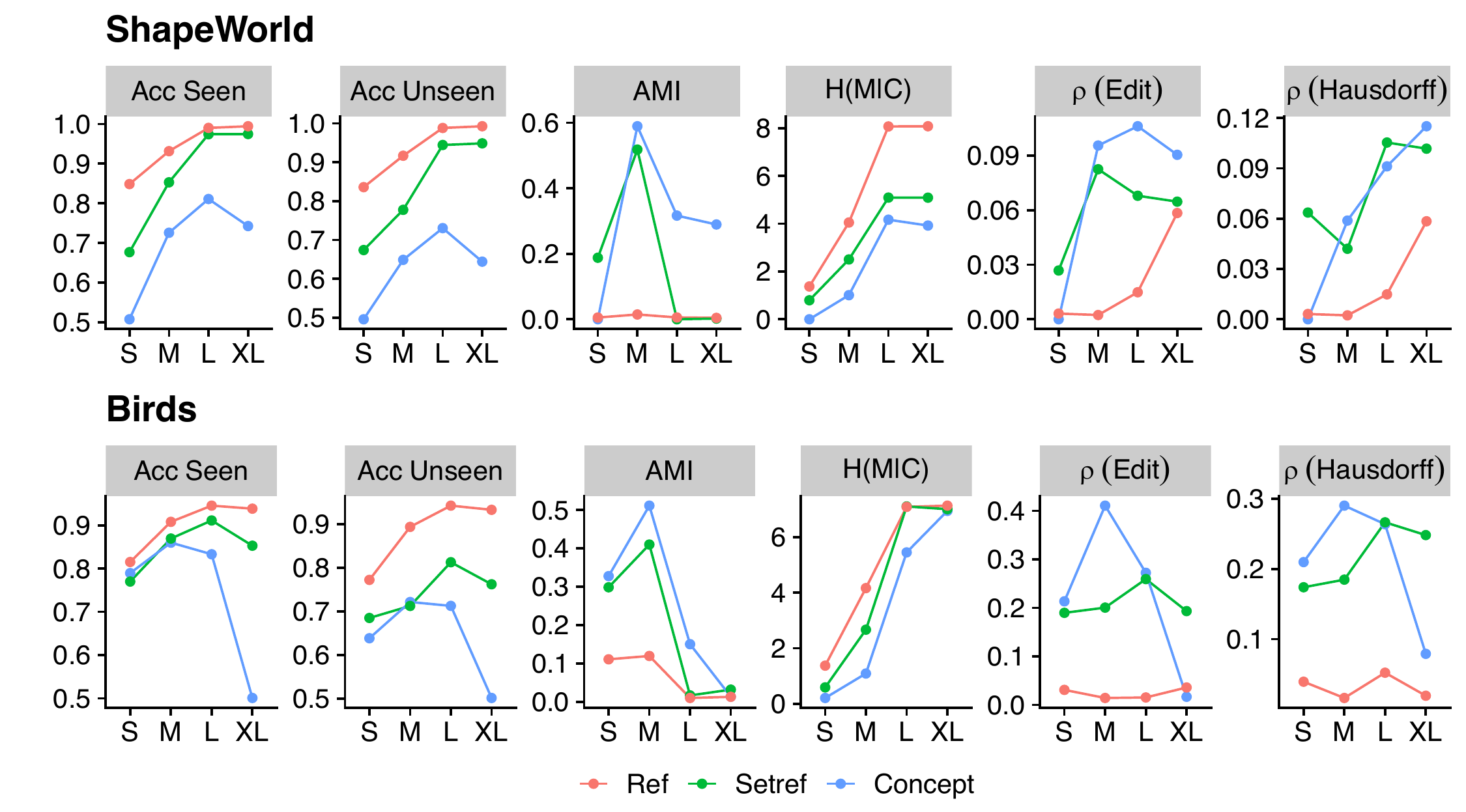}
    \caption{Accuracy and language systematicity metrics while varying the communication channel bandwidth for both datasets. Each dot represents one run.}
    \label{fig:vary_comm}
    \vspace{-1em}
\end{figure}

Results are in Figure~\ref{fig:vary_comm}. Our conclusions are as follows:

\begin{enumerate}
    \item Training for concept games is less stable, and with very small (Shapeworld S) and very large (Birds XL) vocabulary sizes is often unable to learn.
    \item Outside of the concept game agent failures, accuracy on both seen and unseen concepts tends to increase as we increase the communication channel size.
    \item Across all games, the information theoretic measures tend to show less systematicity as we increase the communication bandwidth. This is likely because as the number of vocabulary tokens and message length increases, the chance of sampling an errant token while generating a message increases, which is then treated as a completely unique message under our information theoretic measures; thus some increased entropy and reduced AMI is expected.
        \begin{enumerate}
            \item Regardless, when comparing between games with equal channel sizes, setref and concept games generally have more systematic language. Concept games are more consistently systematic than reference games, except for the degenerate settings where it is unable to learn. Setref games are more systematic than reference games, except as measured by $\AMI$ in large vocabulary spaces (L, XL). The differences across the game types are smaller when the channel size is very large, e.g.\ in Birds L and XL.
        \end{enumerate}
    \item Similarly, across all channel sizes, topographic $\rho$ tends to be higher for setref and concept games. Somewhat mysteriously, the topographic $\rho$ measures do not respond to varying channel sizes as the information theoretic measures do. In fact, for ShapeWorld, increasing channel size \emph{increases} topographic $\rho$, especially for reference games. More work is needed to identify the sources of this effect.
\end{enumerate}

To summarize: for communication channel sizes reasonably sized according to the dataset (e.g.\ the medium setting here, and the setting reported in the main text), our conclusions hold; for extremely small or large communication channel sizes, our conclusions still hold, but care must be taken to ensure stable training in these regimes, especially for concept game agents.

\section{Upper bounds on listener performance for setref and concept games}
\label{app:upperbounds}

One way of calibrating the quality of emergent languages developed by our agents in setref and concept settings is to evaluate a listener on the set classification task given an ``ideal'' language. While we do not have human data generated for these games, we can use proxies to obtain upper bounds on listener performance. For ShapeWorld, we give the listener the ground-truth concept descriptions; for Birds, we use a randomly sampled caption associated with one of the images in the target class, using the language corpus collected by Reed \emph{et al.} \cite{reed2016learning}. Table~\ref{tab:upperbounds} shows results. With the true ShapeWorld concepts, listeners are able to attain near-perfect performance, suggesting that the agents in our settings have an imperfect understanding of the true concepts in each game. However, for Birds performance, emergent language performance is actually comparable to the ground-truth language performance (at 79\% and 71\%) performance on seen and unseen tasks, respectively). This suggests that the emergent languages for this task do quite well, even reaching the upper bounds on performance for the agent architectures we explore in this paper.

\begin{table}[h]
    \centering
    \caption{Performance of listener agents on the ground-truth setref/concept task when given ideal (human) languages. Note setref and concept are the same, since there is no teacher input.}
    \vspace{1em}
    \begin{tabular}{lrr}
    \toprule
     Dataset &  Acc (Seen) & Acc (Unseen) \\
     \midrule
     ShapeWorld &  99.8 (0.1) & 99.8 (0.1) \\
     Birds &  79.3 (0.4) & 70.6 (2.0) \\
     \bottomrule
    \end{tabular}
    \label{tab:upperbounds}
\end{table}

\section{Experiments with traditional cross-entropy reference games}
\label{app:refgame}

We presented an atypical formulation of reference games as consisting of multiple targets, with student decisions made independently:
\begin{align}
p^S(Y^S \mid X^S, m) = \prod_{i} p^S(y^S_i \mid x^S_i, m),
\end{align}
where students are trained with the binary cross entropy loss in Equation~\ref{eq:bce}, restated here for convenience:
\begin{align}
    \mathcal{L}_{\text{BCE}}(T, S, G) = - \sum_i \log p^S(y^S_i \mid x^S_i, \hat{m}), \quad \hat{m} \sim p^T(m \mid X^T, Y^T).
\end{align}
This was done to keep training objectives and models identical across games, and to keep the amount of training data consistent (i.e.\ there are exactly the same number of targets and distractors seen by each agent across training).

However, the typical reference game has a single target: instead of $Y^S \in \{0, 1\}^n$, we have a single target $t^S \in [1, n]$ denoting the index of the single positive example. Then the student probability that input $i$ is the target is the softmax-normalized
\begin{align}
p^S(i \mid X^S, m) = 
\frac{\exp( \textsc{RNN-Encode}(m) \cdot f^S_\phi(x^S_i) )}{\sum_{i'} \exp( \textsc{RNN-Encode}(m) \cdot f^S_\phi(x^S_{i'}) )} \nonumber
\end{align}
and the training objective for a single game is
\begin{align}
\mathcal{L}_{\textsc{XENT}}(T, S, G) = - \log p^S(t^S \mid x^S_i, \hat{m}), \quad \hat{m} \sim p^T(m \mid X^T, Y^T).
\end{align}

\begin{figure*}[b]
    \centering
    \includegraphics[width=\textwidth]{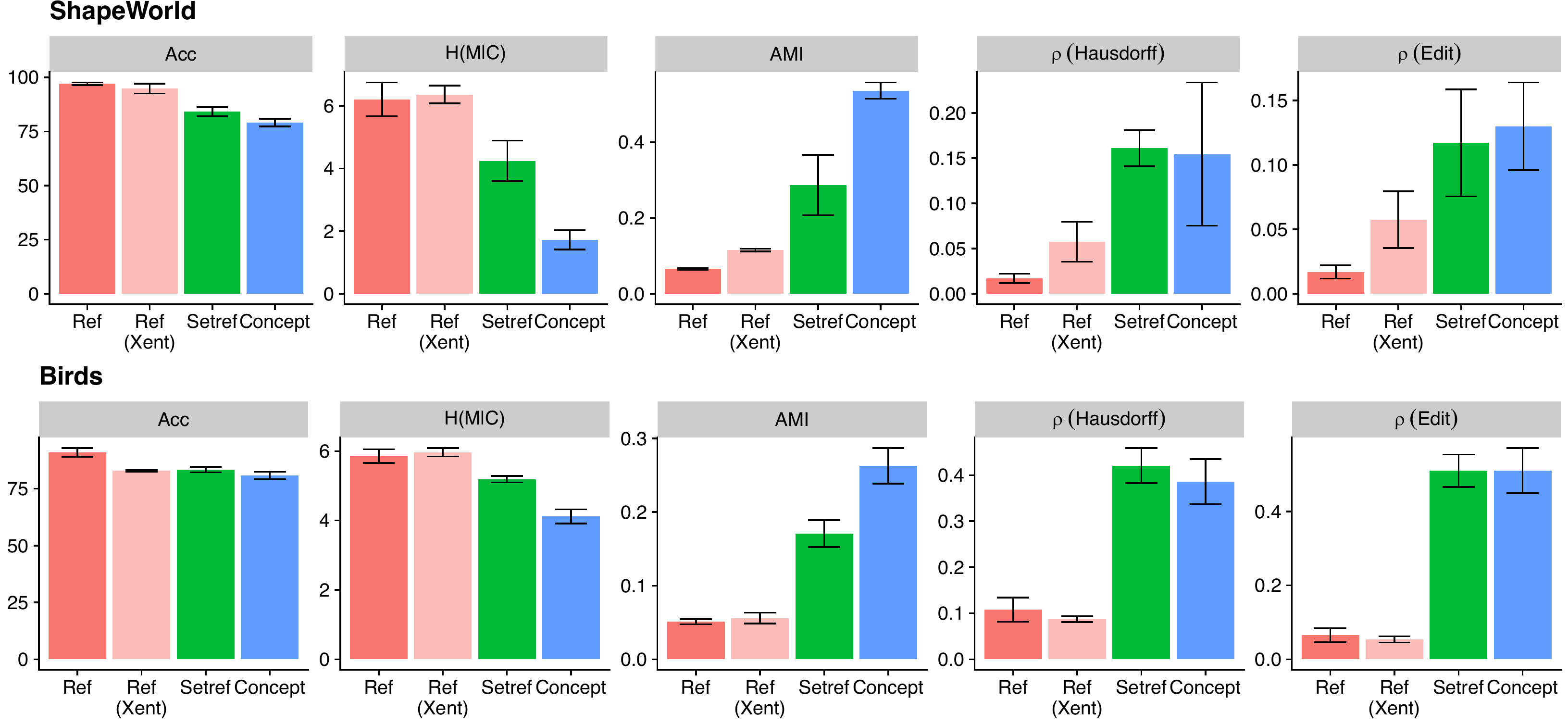}
    \caption{Accuracy and measures of language systematicity for reference games, setref games, and concept games, as well as reference games trained with the traditional cross entropy (xent) objective.}
    \label{fig:xent}
    \vspace{-1em}
\end{figure*}

To ensure that our alternative formulation did not affect results, we ran 5 experiments with the standard reference game trained with cross entropy loss, with a single target and 10 distractors. Figure~\ref{fig:xent} summarizes the relevant statistics; besides slightly higher topographic $\rho$ and $\AMI$ for the cross-entropy reference games for ShapeWorld, there are no qualitative differences compared to our reference game formulation and our conclusions are unchanged.

\section{Additional plots of speaker messages}
\label{app:sunburst}

\begin{figure*}
    \centering
    \includegraphics[width=\textwidth]{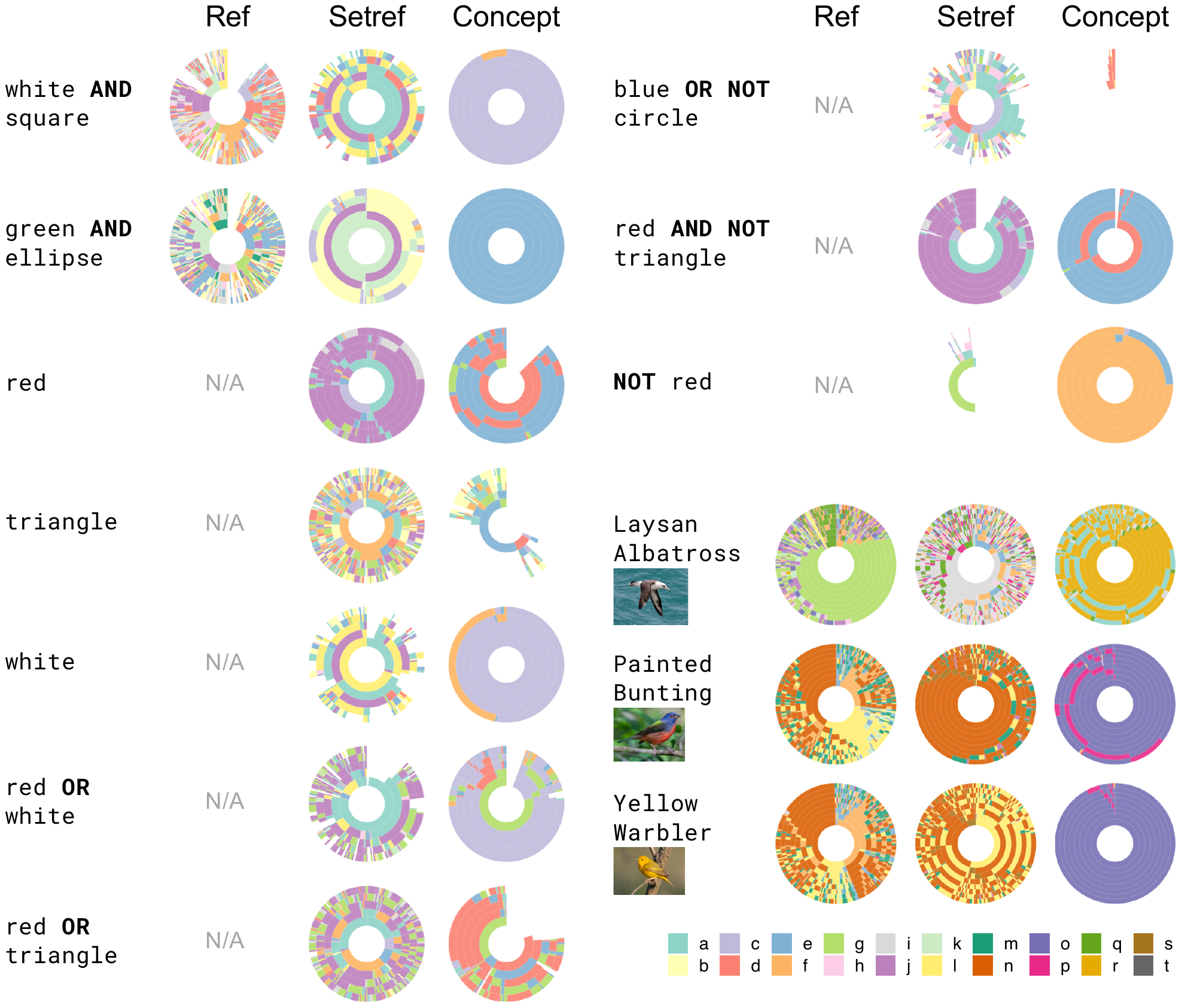}
    \caption{Additional plots of teacher messages for selected ShapeWorld and Birds games. Most ShapeWorld concepts are not tested in reference games, so those plots are not available.}
    \label{fig:extra_sunburst}
    \vspace{-1em}
\end{figure*}

See Figure~\ref{fig:extra_sunburst} for additional plots of teacher messages made for ShapeWorld and Birds games. Overall, the plots show a general reduction in language complexity from ref to setref to concept, although some quirks emerge: for example, some characters (e.g.\ \emph{e} in concept) appear to be overloaded (across \emph{green ellipse} and \emph{red}), and concept uses similar language for \emph{painted bunting} and \emph{yellow warbler}. White gaps indicate end of sentence, so there are games where the speaker teacher utters nothing (e.g.\ \emph{blue or not circle} concept; \emph{not red} setref).

\section{ACRe model, training, and dataset details}
\label{app:acredetails}

Like the agents, ACRe models are trained with 
the Adam optimizer and learning rate 0.0001 on a Titan Xp 
GPU. Training ACRe takes around 1-2 hours on 4 CPUs and 16GB of RAM.

\paragraph{Models.}
$\hat{p}^c_\eta$ is implemented as an unconditional Transformer \cite{vaswani2017attention} LM with 2 hidden layers, 2 attention heads, a word embedding size of 50, hidden and intermediate sizes of 100, and $\hat{p}^\omega_\eta$ has the same decoder, but also has a Transformer encoder with identical parameters and cross attention from the decoder to the encoder. The vocabulary of the Transformers are the vocabulary of the emergent language plus a special \texttt{[SEP]} token used to concatenate argments for higher-order $\omega$ operations. These are implemented with Huggingface's Transformers library \cite{wolf2020transformers}.

Concretely, this means that for ShapeWorld, we have 11 unconditional transformer LMs modeling each $\hat{p}^c_\eta$, and 3 transformer LMs, one modeling the unary operation NOT, and two modeling AND and OR.

\paragraph{Data.} To train ACRe to approximate the language for a teacher-student pair, we sample 200000 messages from the teacher, evenly distributed across all games. These are then stratified into data used to train each primitive LM and data used to train each higher order operation. All LMs are trained with standard language modeling techniques with teacher forcing. We first train the primitive LMs on their respective data. Then we train the unary NOT model on concepts of the form $\text{NOT}(c)$ where $c$ is primitive, sampling a message for $c$ using the primitive models. Finally, we train the AND and OR models on the conjunctive and disjunctive concepts, sampling arguments from the primitive models---and the NOT model, if the argument is negated.

We train ACRe models for 20 epochs and do early stopping as follows:
for the training of the primitives and the NOT model, we divide the agent messages into 90\% training data and 10\% data, and define one epoch as one pass through the training data, stopping training once performance on the validation data is maximized. For the training of the AND and OR models, since we would like to test compositional generalization, we divide the data \emph{by concept}: we stratify all unique conjunctions and disjunctions into a train/val/test split of 80\%, 10\%, and 10\% respectively. Models are trained on messages representing the 80\% of training concepts (1 epoch = 1 pass through this data), with early stopping performed when performance is maximized on the messages generated for the unseen validation concepts. Finally, we evaluate on the final 10\% of unseen test concepts. To produce the numbers presented in Table~\ref{tab:acre}, we evaluate listener and BLEU-1 performance across 110000 ShapeWorld games and 7000 Birds games (i.e.\ 5 passes through the train and test data for each dataset), grouping the metrics by whether the concept belongs to either the train/val or test ACRe splits.

Note that we do not test generalization of the NOT operation as there are too few NOT operations to be able to generalize properly. We verify this by training our NOT model to attempt to predict the (perfectly compositional) \emph{ground truth concepts}: in other words, the task is to predict \emph{green} from \emph{not green}, \emph{blue} from \emph{not blue}, and then use these to finally predict the negated version of \emph{red}. However, the model completely fails to generalize as there are only 10 unique concepts, so the model simply memorizes the training set. Note that we still use the NOT model in sampling messages for conjunctive and disjunctive concepts that include $\text{NOT}(c)$ as an argument.

\end{document}